\documentclass[letterpaper]{article}
\usepackage[preprint]{aaai2027}

\usepackage{amsmath}
\usepackage{amsthm}
\usepackage{amssymb}
\usepackage[most]{tcolorbox}
\usepackage{booktabs}
\usepackage{colortbl}
\usepackage{array}
\usepackage{xcolor}
\usepackage{enumitem}
\usepackage{pifont}
\usepackage[hyphens]{url}
\usepackage{graphicx}
\usepackage{natbib}
\usepackage{caption}
\usepackage{algorithm}
\usepackage{algorithmic}
\usepackage{newfloat}
\usepackage{listings}

\definecolor{cobaltdark}{RGB}{25,70,140}
\definecolor{cobaltlight}{RGB}{235,243,255}
\definecolor{jsonbg}{RGB}{240,246,255}
\definecolor{rowgray}{RGB}{244,247,251}
\definecolor{okgreen}{RGB}{0,120,60}
\definecolor{badred}{RGB}{190,30,45}

\newtcolorbox{promptbox}[1]{%
  enhanced, breakable,
  colback=cobaltlight, colframe=cobaltdark,
  fonttitle=\bfseries\small, coltitle=white,
  title={#1}, boxrule=0.6pt, arc=2pt,
  left=7pt, right=7pt, top=5pt, bottom=5pt,
  before skip=8pt, after skip=8pt}

\lstdefinestyle{jsonstyle}{%
  backgroundcolor=\color{jsonbg}, basicstyle=\ttfamily\footnotesize,
  breaklines=true, frame=single, rulecolor=\color{cobaltdark!40},
  xleftmargin=6pt, xrightmargin=6pt, framesep=4pt, showstringspaces=false}

\DeclareCaptionStyle{ruled}{labelfont=normalfont,labelsep=colon,strut=off}
\floatstyle{ruled}
\newfloat{listing}{tb}{lst}{}
\floatname{listing}{Listing}

\newcommand{\yes}{\textcolor{okgreen}{\checkmark}}
\newcommand{\no}{\textcolor{badred}{\ding{55}}}
\newcommand{\src}[1]{\small\textsf{#1}}
\newcommand{\samethanks}{\footnotemark[\value{footnote}]}

\theoremstyle{definition}
\newtheorem{definition}{Definition}

\title{\textsc{Genesis}: Towards Explainable Causal Discovery}

\author{
    Abhinav Thorat,
    Ravi Kumar Kolla,
    Vishak K Bhat\thanks{Work done during an internship at Sony Research India.},
    Harsh Vardhan Singh Chauhan\samethanks,\\
    Niranjan Pedanekar
}
\affiliations{
    User Engagement Research, Sony Research India\\
    \{abhinav.thorat, ravi.kolla, vishak.bhat, harsh.chauhan1,\\
    niranjan.pedanekar\}@sony.com
}

\begin{document}
\maketitle

\begin{abstract}
Causal Discovery (CD) from observational data faces two fundamental challenges: statistical methods lack the power to reliably resolve structural ambiguities in low-sample regimes, while LLM-assisted hybrid approaches improve causal structure recovery by incorporating semantic reasoning whose influence on individual edge decisions remains opaque. As a result, existing hybrid methods largely overlook a fundamental requirement: explaining why a particular edge is included or excluded in the learned directed acyclic graph (DAG). This limitation is especially critical in real-world applications where no ground-truth DAG exists and every structural decision must be independently justified. We formalize this requirement as \emph{decision traceability}, requiring every inferred edge to be supported by auditable statistical evidence, Markov Blanket consistency, or explicit domain reasoning. We propose \textsc{Genesis}, an explainable hybrid CD framework that decomposes graph construction into interpretable decision points. \textsc{Genesis} first identifies and scores three-node structural motifs (chains, forks, and colliders) to establish transparent structural priors before progressively integrating them with observational evidence through a refinement process that invokes domain knowledge only when statistical evidence is insufficient. By construction, \textsc{Genesis} guarantees decision traceability by ensuring that every edge decision is resolved through auditable evidence sources. Experimental results confirm that \textsc{Genesis} achieves 100\% decision traceability across all experimental settings, establishing explainability as a first-class objective in CD. Despite this additional requirement, \textsc{Genesis} consistently outperforms statistical CD methods on the majority of benchmark datasets across all sample regimes in terms of Structural Hamming Distance (SHD), while achieving performance comparable to state-of-the-art LLM-assisted approaches.
\end{abstract}

\section{Introduction}
\label{sec:introduction}
Causal Discovery (CD) aims to recover causal relationships among variables from observational data, providing the structural foundation for interventional reasoning, counterfactual analysis, and causal decision-making~\cite{pearl2009causality, peters2017elements}. This capability is beneficial for downstream applications including treatment effect estimation \cite{gupta2023local}, root cause analysis \cite{ikram2022root}, recommendation systems \cite{zheng2021disentangling}, and scientific discovery \cite{sachs2005causal}. However, recovering causal structures from observational data remains challenging due to limited samples, statistical noise, and latent confounders \cite{glymour2019review,colombo2012learning}. A further obstacle is Markov equivalence, where multiple directed acyclic graphs (DAGs) induce identical observational distributions~\cite{spirtes2000causation,pearl2009causality}.

While Structural Hamming Distance (SHD) remains the standard metric for evaluating CD algorithms, it is meaningful only when a ground-truth DAG is available and offers no insight into why individual edge decisions are made. This limitation is particularly significant in real-world applications, where ground-truth causal graphs are rarely available. In such settings, an equally important yet largely overlooked objective is \emph{decision traceability}: the ability to justify why each edge is included or excluded in the learned DAG through explicit, auditable evidence.
 
Existing statistical (traditional) CD methods optimize global statistical objectives or local independence criteria without providing transparent reasoning for individual edge decisions. Constraint-based methods such as PC~\cite{spirtes2000causation} apply conditional independence tests iteratively but lack explicit justifications for why specific tests succeed or fail. Score-based methods such as NOTEARS~\cite{zheng2018dags} optimize continuous relaxations of combinatorial objectives, making edge-level decisions difficult to interpret. Identifiable functional approaches including LiNGAM~\cite{shimizu2006linear}, DirectLiNGAM~\cite{shimizu2011directlingam}, SCORE~\cite{rolland2022score}, and additive noise models (ANM)~\cite{hoyer2008nonlinear} recover causal directions under specific structural and distributional assumptions, yet provide limited transparency into the evidence underlying individual edge orientations.  While these methods have demonstrated success across domains, they struggle in low-sample regimes and produce graphs whose edge-level decisions are difficult to audit. This limitation is particularly important for downstream applications such as root cause analysis, recommendation systems, and scientific discovery, where causal analyses and interventions benefit from accurate, interpretable, and auditable causal structures.

Recent advances in large language models (LLMs) offer a complementary paradigm for CD. Through exposure to large-scale textual corpora, LLMs encode semantic relationships, domain priors, and commonsense causal patterns that may not be directly observable from limited observational data alone. 
Recent works~\cite{vashishtha2025causal, ban2025integrating, du2025causal} have explored the use of LLMs for CD, showing that semantic priors extracted from variable metadata can complement statistical structure learning and improve causal graph recovery in many settings.
However, existing LLM-assisted approaches tightly couple semantic reasoning with statistical evidence without explicitly tracing how each contributes to the final graph. Consequently, they provide little insight into why a particular edge is included or excluded, making it difficult to determine whether a decision is supported by observational evidence, semantic priors, or implicit knowledge acquired during pretraining. This lack of decision traceability limits the interpretability and auditability of learned causal structures.

We address these limitations through \textsc{Genesis}, a hybrid framework that integrates motif-level LLM reasoning with statistical validation while enforcing \emph{decision traceability}  by design. We define decision traceability as the property that every edge in the learned DAG is associated with an explicit, auditable justification grounded in one or more of: (i) statistical evidence computed from the observational data; (ii) Markov Blanket (MB) consistency between data-driven and semantic inference; or (iii) explicit domain reasoning via counterfactual-style LLM queries. Rather than treating explainability as a post-hoc objective, \textsc{Genesis} embeds it directly into structure learning. Instead of reasoning over isolated edges, our framework operates on \emph{3-node causal motifs}, chains, forks, and colliders~\cite{pearl2009causality}, which capture identifiable conditional independence patterns through $d-$separation and provide a principled representation for integrating semantic priors with statistical evidence.

\textsc{Genesis} comprises three stages. An \emph{LLM Explorer} first extracts candidate motifs with confidence scores from node metadata to establish structural priors. A \emph{Data Explorer} then evaluates motif- and edge-level plausibility through complementary statistical evidence, including conditional independence tests, BIC scoring~\cite{schwarz1978estimating}, additive noise models~\cite{hoyer2008nonlinear}, and LiNGAM~\cite{shimizu2006linear}. The resulting semantic and statistical confidence scores are combined to construct an initial graph. This graph is subsequently refined through \emph{MB filtering}~\cite{tsamardinos2003algorithms}, extending it by intersecting data-driven and LLM-derived blankets, and only when necessary through a \emph{Judge LLM} that performs counterfactual reasoning to resolve residual edge ambiguities. This progressive refinement strategy incrementally resolves structural ambiguities, preserving the complementary strengths of semantic and statistical evidence while ensuring that every graph construction decision remains explicitly traceable.

We evaluate \textsc{Genesis} on five benchmark Bayesian networks (Asia, Cancer, Child, Earthquake, Survey)~\cite{scutari2010learning} across sample sizes ranging from 1000 to 5000, comparing against traditional methods such as PC, SCORE, NOTEARS, CaMML, DirectLiNGAM, and recent LLM-assisted approaches. Experimental results confirm that \textsc{Genesis} achieves 100\% decision traceability across all experimental settings. Despite this additional requirement, \textsc{Genesis} consistently outperforms statistical CD methods on the majority of benchmark datasets across all sample regimes in terms of SHD, while achieving performance comparable to LLM-assisted approaches. \\
\textbf{Contributions.} We summarize our contributions as follows:
\begin{itemize}
\item We formalize \emph{decision traceability} as a
complementary objective to SHD for CD, requiring every learned
edge to be supported by explicit, auditable evidence.

\item We introduce 3-node causal motifs (chains, forks, and colliders) as the
unit of reasoning in place of isolated edges, providing an intermediate
representation that encodes identifiable $d$-separation patterns and imposes
stronger structural constraints than pairwise edge prediction.

\item We propose \textsc{Genesis}, a hybrid CD framework that
resolves each edge through a progressive cascade of statistical validation,
MB consistency, and counterfactual LLM arbitration, invoking each
stage only when the preceding one leaves the decision unresolved.

\item We demonstrate on benchmark Bayesian networks that \textsc{Genesis}
achieves 100\% decision traceability across all experimental settings while
consistently outperforming statistical CD methods on the
majority of datasets and achieving performance comparable to state-of-the-art (SOTA) LLM-assisted approaches in terms of SHD.
\end{itemize}
\section{Literature Survey}
\label{sec:literature-survey}
\textbf{Traditional CD.} These methods include constraint-based, score-based, and identifiable functional approaches. Constraint-based methods (e.g., PC~\cite{spirtes2000causation}) infer causal structure through conditional independence tests, score-based methods (e.g., CaMML~\cite{wallace1996causal} and NOTEARS~\cite{zheng2018dags}) optimize objectives over candidate DAGs, while identifiable functional models (e.g., LiNGAM~\cite{shimizu2006linear}, DirectLiNGAM~\cite{shimizu2011directlingam}, and SCORE~\cite{rolland2022score}) recover causal directions under specific structural and distributional assumptions. Although effective, these methods primarily optimize graph recovery and provide limited insight into \emph{why} particular edge decisions are made. Consequently, edge orientations are driven by statistical tests, optimization objectives, or modeling assumptions without explicit, auditable justification. In contrast, \textsc{Genesis} performs causal discovery through interpretable three-node motifs, progressively resolving uncertain edges using statistical evidence, consensus MBs, and, only when necessary, LLM-based reasoning to produce a decision-traceable DAG. \\
\textbf{LLM-based CD.} Recent work has explored the use of LLMs for CD by leveraging semantic and domain knowledge. The authors in~\cite{vashishtha2025causal} infer causal orderings among variables to constrain downstream structure learning, while approaches such as LLM-CD~\cite{ban2025integrating,du2025causal} predict causal edges or graph structures from variable descriptions and use them as priors for subsequent CD.
Although these methods demonstrate the value of LLM-derived priors, they primarily rely on pairwise edge predictions or global ordering assumptions, making them susceptible to ambiguities from Markov equivalence and conflicts between semantic priors and observational data. In contrast, \textsc{Genesis} generates \emph{motif-level} hypotheses that are progressively validated against observational evidence, ensuring every retained edge is supported by both semantic reasoning and empirical evidence, yielding an auditable and decision-traceable CD process.

\begin{algorithm}
\caption{\textsc{\textsc{Genesis:} LLM+Data Guided Causal Discovery via Motifs}}
\label{alg:proposed-algorithm}
\begin{algorithmic}[1]
\STATE \textbf{Input:} Observational data $D$, node metadata $M$, $\mathrm{LLM_{mf}},$ $\mathrm{LLM_{mb}},$ $\mathrm{LLM_{hyp}},$ $\mathrm{LLM_{judge}},$ hyperparameters 
$\{w_1, w_2, w_3, w_4, \alpha, \beta_0, \beta_1, \mathtt{\lambda_{low}}, \mathtt{\lambda_{high}}\}$
\STATE \textbf{Output:} DAG $G$
\STATE \textbf{\textsc{Step 1: LLM Explorer}}
\STATE Query the $\mathrm{LLM_{mf}}$ using node metadata $M$ to extract 3-node motifs (chain, fork, collider) along with confidence scores in $[0,1]$.
\STATE Let $\mathcal{M}$ denote the set of motifs outputted by the $\mathrm{LLM_{mf}}$, and $p(m \mid \mathrm{LLM_{mf}})$ denote the confidence for $m \in \mathcal{M}$.
\STATE \textbf{\textsc{Step 2: Data Explorer}}
\STATE \hspace{0.5em}\textit{(a) Edge confidence}
\STATE Extract unique edges $E_1$ from $\mathcal{M}$.
\STATE For each $e \in E_1$, compute edge confidence using data $D$ as follows:
\begin{multline*}
p(e \mid D) = w_1\,p(e \mid D, \mathrm{CI}) + w_2\,p(e \mid D, \mathrm{BIC}) + \\
w_3\,p(e \mid D, \mathrm{ANM}) + w_4\,p(e \mid D, \mathrm{LiNGAM}) 
\end{multline*}
\STATE \hspace{0.5em}\textit{(b) Motif confidence}
\STATE For each $m \in \mathcal{M}$, compute $p(m \mid D)$ same as using CI, BIC, ANM and LiNGAM by verifying motif-specific conditions.
\STATE \textbf{\textsc{Step 3: Edge Resolution}}
\FOR{each motif $m \in \mathcal M$}
    \FOR{each edge $e \in m$}
        \STATE $s(e,m) \gets \alpha\, p(m \mid \mathrm{LLM_{mf}}) + (1-\alpha)\big(\beta_0\, p(e \mid D) + \beta_1\, p(m \mid D)\big)$
    \ENDFOR
\ENDFOR
\STATE Initialize $
G=\{e:\exists\,m\in\mathcal M,\ s(e,m)\ge\lambda_{\text{low}}\}.$
\STATE Define the uncertain edges set as CF edges, $E_2$:
\[
E_2:=\left\{
e\in G \; \middle| \;
\begin{array}{l}
\text{(i) } s(e,m)\in[\lambda_{\text{low}},\,\lambda_{\text{high}}],\\[0.3em]
\text{\hspace{1.8em}or}\\[0.3em]
\text{(ii) } p(m|\mathrm{LLM}_{\mathrm{mf}})\, \text{and}\,
p(m|D)\, \text{are} \\
\hspace{1.5em}\text{inconsistent for at least one } \\
\hspace{1.5em} \text{motif }m\ni e.
\end{array}
\right\}.
\]
\FOR{each edge $e: a \rightarrow b \in E_2$}
\STATE \textit{(a) Markov Blanket filtering}
\STATE Compute $\mathrm{MB}(a \mid D),$ $\mathrm{MB}(a \mid \mathrm{LLM_{mb}}),$ $\mathrm{MB}(b \mid D)$ and $\mathrm{MB}(b \mid \mathrm{LLM_{mb}})$ for both nodes $a$ and $b$ in the edge $e$, then define:
\[
\mathrm{MB}(a) := \mathrm{MB}(a \mid D) \cap \mathrm{MB}(a \mid \mathrm{LLM_{mb}})
\]
\[
\mathrm{MB}(b) := \mathrm{MB}(b \mid D) \cap \mathrm{MB}(b \mid \mathrm{LLM_{mb}})
\]
\IF{$a \in \mathrm{MB}(b)$ \AND $b \in \mathrm{MB}(a)$}
\STATE \texttt{KEEP} edge $e$ in $G$
\ELSE
\STATE \textit{(b) Counterfactual LLM resolution}
\STATE Generate hypothesis set $\mathcal{H}(e) \leftarrow
       \mathrm{LLM}_{\mathrm{hyp}}(e, M)$: five competing structural
       explanations for $e$ over $M$.
\STATE Query $\mathrm{LLM}_{\mathrm{judge}}$ with $(e, M, \mathcal{H}(e))$
       using counterfactual-style questioning over the alternatives in
       $\mathcal{H}(e)$.
\STATE KEEP $e$ in $G$ if output is KEEP, else DISCARD.
\ENDIF
\ENDFOR
\STATE \textbf{Return} $G$
\end{algorithmic}
\end{algorithm}

\section{Proposed Model: \textsc{Genesis}}
\label{sec:proposed-model} 
\begin{figure*}[t]
\centering
\includegraphics[width=\textwidth]{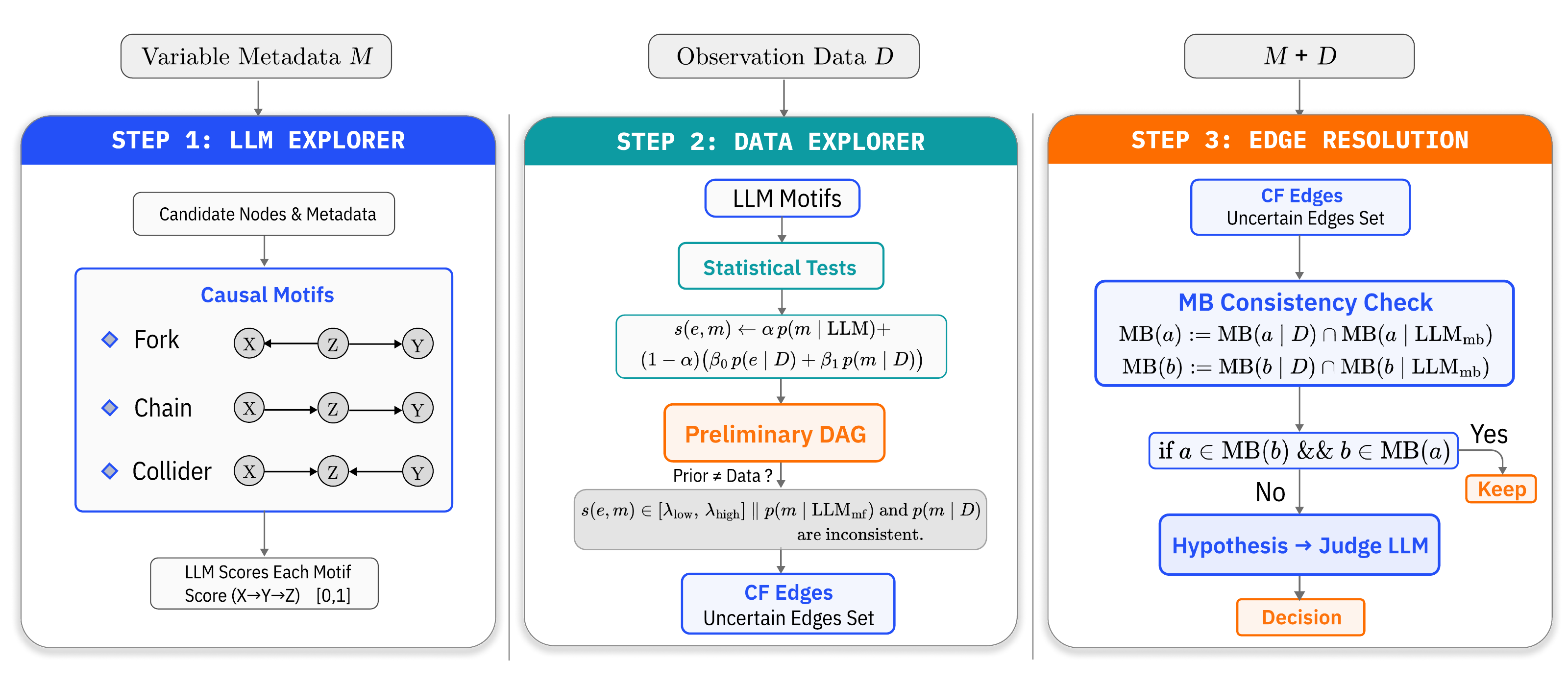}
\caption{
\textbf{GENESIS architecture.} The framework consists of three stages:  \textbf{(1) LLM Explorer} extracts candidate motifs with semantic confidence;  \textbf{(2) Data Explorer} estimates edge and motif confidence from the observational data to construct an initial DAG; and  \textbf{(3) Edge Resolution} refines uncertain edges using consensus MB filtering and, when required, a Judge LLM, producing a decision-traceable DAG.
}
\label{fig:architecture}
\end{figure*}

\begin{figure*}[t]
    \centering
    \includegraphics[width=\linewidth]{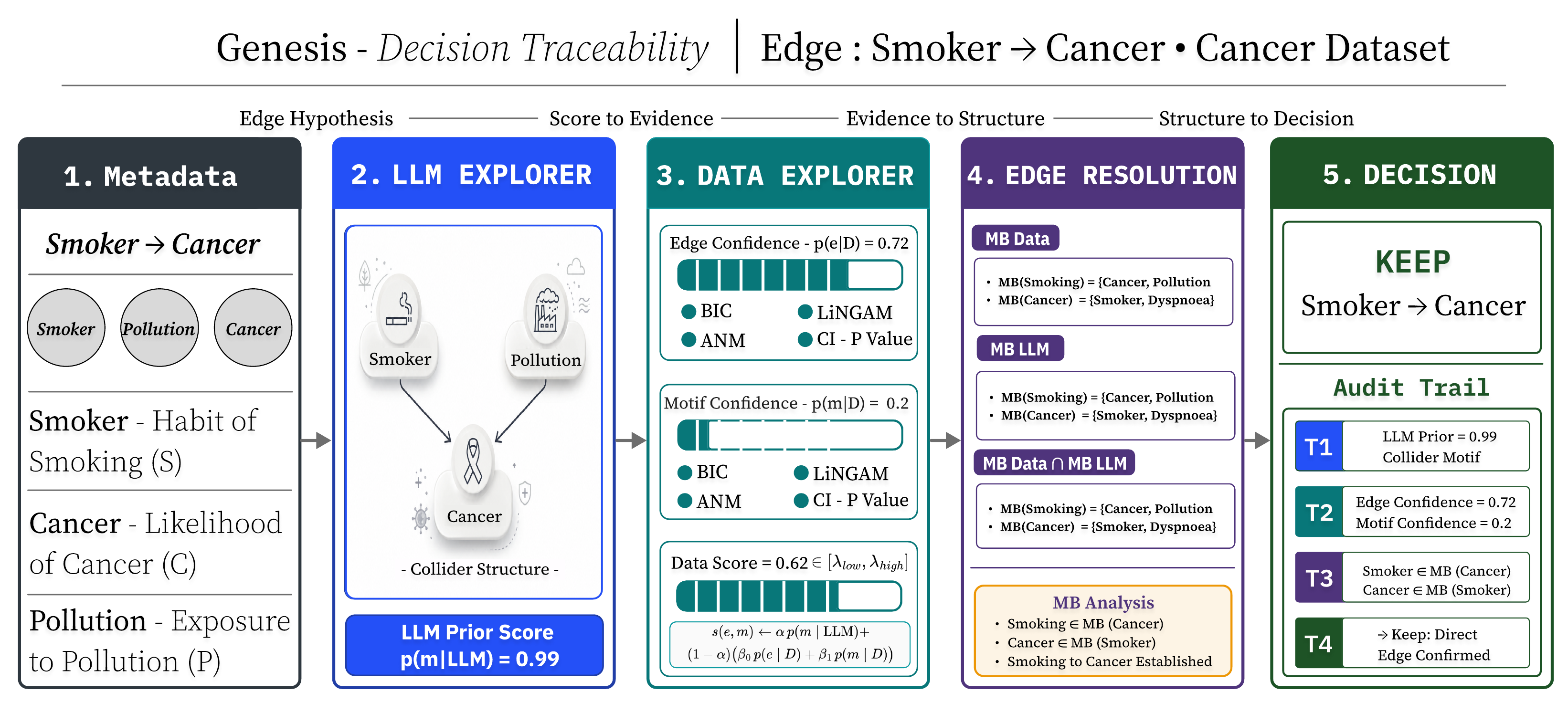}
    \caption{Decision traceability trace for the edge \textit{Smoker} 
    $\rightarrow$ \textit{Cancer} in the Cancer Bayesian Network. The pipeline 
    flows left to right: LLM prior scoring over structural motifs establishes 
    an interpretable prior, statistical validation quantifies data-driven 
    evidence, and MB resolution across all three \textbf{GENESIS} 
    variations determines the final edge decision. Each stage produces an 
    auditable justification traceable to motif plausibility, conditional 
    independence evidence, or MB membership, collectively 
    forming a complete audit trail for every structural decision.}
    \label{fig:traceability}
\end{figure*}

In this section, we present \textsc{Genesis}, illustrated in Figure~\ref{fig:architecture}, with the complete procedure summarized in Algorithm~\ref{alg:proposed-algorithm}. \textsc{Genesis} is a hybrid CD framework that integrates LLM-derived semantic priors with statistical evidence using \emph{3-node causal motifs}. Using motifs as the fundamental reasoning units reduces structural ambiguity while exploiting identifiable structural constraints beyond pairwise reasoning. 

Given observational data $D$ and node metadata $M$, the objective is to recover a DAG $G$ together with explicit justifications for every inferred edge, formalized below as \emph{decision traceability}. \textsc{Genesis} achieves this through a progressive three-stage pipeline comprising (i) \textbf{LLM Explorer}, which generates candidate 3-node motifs from node metadata; (ii) \textbf{Data Explorer}, which validates these motifs using observational data; and (iii) \textbf{Edge Resolution}, which resolves structural ambiguities through MB consistency and counterfactual Judge LLM reasoning.
\begin{definition}[Decision Traceability]
\label{def:decision-traceability}
A CD framework satisfies \textit{decision 
traceability} if, for every edge $e$ in the learned DAG $G$, 
there exists an explicit, auditable justification $J(e)$ drawn 
from one or more of the following evidence sources:
\begin{enumerate}
    \item \textbf{Statistical evidence}: Each edge $e$ is supported by statistical CD methods such as
    conditional independence tests, score-based model fit, or 
    functional causal model estimates computed from 
    observational data $D$, detailed in Data Explorer Section.
    \item \textbf{Markov Blanket (MB) consistency}: Both nodes  
    of an edge $e$ appear in each other's validated MB, 
    confirming local structural necessity from both data-driven 
    and LLM-derived neighborhood estimates, detailed in the Edge Resolution Section.
    \item \textbf{Domain reasoning}: a Judge LLM provides an 
    explicit causal hypothesis grounding edge $e$ in mechanistic 
    domain knowledge, invoked only when statistical evidence 
    and MB validation are insufficient, detailed in Edge Resolution Section.
\end{enumerate}
\end{definition}

\emph{Remark.} \textsc{Genesis} is designed to satisfy decision traceability by construction: every edge decision satisfies at least one of the three evidence sources above before being included in the final DAG.
\subsection{LLM Explorer: Motif-Level Hypothesis Generation}
\label{sec:llm-explorer}
\textbf{Motivation.} Inferring causal relations between pairs of variables from observational data is inherently ambiguous due to Markov equivalence and limited edge orientation information. In contrast, 3-node motifs capture local causal structures together with their associated conditional independence patterns, providing richer structural constraints than isolated pairwise relations. This richer representation makes motifs a more informative unit for integrating LLM-derived semantic priors with statistical evidence. \\
\begin{table}[t]
\centering
\caption{Distribution of decision justification sources across datasets ($n=1000$). Values denote the percentage of final DAG edges supported by each evidence source. All edges are decision-traceable by construction.}
\resizebox{\columnwidth}{!}{%
\begin{tabular}{lcccc}
\toprule
\textbf{Dataset} & \textbf{Statistical} & 
\textbf{MB Consistency} & \textbf{Judge LLM} & 
\textbf{Decision Traceability} \\
\midrule
Earthquake & 0.00\%  & 100.00\% & 0.00\%  & 100\% \\
Cancer     & 25.00\% & 75.00\%  & 0.00\%  & 100\% \\
Survey     & 0.00\%  & 100.00\% & 0.00\%  & 100\% \\
Asia       & 0.00\%  & 77.78\%  & 22.22\% & 100\% \\
Child      & 42.31\% & 46.15\%  & 11.54\% & 100\% \\
\bottomrule
\end{tabular}}
\label{tab:traceability}
\end{table}
\textbf{Motif Extraction.} We query an LLM, denoted by $\mathrm{LLM_{mf}},$ using node metadata $M$ to generate candidate 3-node motifs of the following types: chain ($X \rightarrow Y \rightarrow Z$), fork ($X \leftarrow Y \rightarrow Z$), and collider ($X \rightarrow Y \leftarrow Z$). Each motif $m$ is associated with a confidence score $p(m \mid \mathrm{LLM_{mf}}) \in [0,1]$~\footnote{To control computational complexity, we limit the total number of extracted motifs to at most $5n$, where $n$ is the number of nodes in the graph.}. The resulting set is denoted by:
$\mathcal{M} = \{(m, \, p(m \mid \mathrm{LLM_{mf}}))\}.$
\subsection{Data Explorer: Statistical Validation}
\label{sec:data-explorer}
\textbf{Motivation.} LLM-generated motifs encode prior or semantic knowledge but are not grounded in the observational data. We therefore evaluate both the motifs and their constituent edges using statistical evidence derived from the observational data~$D$. \\
\begin{table}[t]
\centering
\caption{Cancer Dataset (4 edges). Per-evidence accuracy:
\src{Statistical} 1/1 ($100\%$), \src{MB consistency} 3/3 ($100\%$).}
\begin{tabular}{lcc}
\toprule
\textbf{Edge} & \textbf{Resolved by} & \textbf{Correct} \\
\midrule
Cancer $\to$ Xray      & \src{MB consistency}   & \yes \\
Pollution $\to$ Cancer & \src{MB consistency}   & \yes \\
Smoker $\to$ Cancer    & \src{MB consistency}   & \yes \\
Cancer $\to$ Dyspnoea  & \src{Statistical} & \yes \\
\bottomrule
\end{tabular}
\label{tab:trace-cancer}
\end{table}
\textbf{Edge Confidence.}
Let $E_1$ denote the set of unique edges present in $\mathcal{M}$. For each edge $e \in E_1$, we compute a data-driven confidence $p(e \mid D) \in [0, 1]$ as follows:
\begin{multline}
\label{eq:edge-confidence-data}
p(e \mid D) = w_1\,p(e \mid D, \mathrm{CI}) + w_2\,p(e \mid D, \mathrm{BIC}) \\ 
+ w_3\,p(e \mid D, \mathrm{ANM}) + w_4\,p(e \mid D, \mathrm{LiNGAM}),
\end{multline}
where $p(e \mid D, \cdot)$ denotes the edge confidence estimated by the corresponding algorithm using the observational data $D$, hyperparameters $w_i \geq 0$ for all $i$, and $\sum_i w_i = 1$. The selected algorithms capture complementary statistical signals, including conditional independence (CI), score-based model fit (BIC), and functional asymmetries (ANM and LiNGAM). Their weighted aggregation provides a robust estimate of edge confidence for validating LLM-generated motifs against the observational data. \\
\textbf{Motif Confidence.}
For each motif $m \in \mathcal{M}$, we compute a data-driven confidence $p(m \mid D) \in [0,1]$ by evaluating whether the conditional independence constraints implied by the motif are supported by the observational data $D$. Specifically, we assess whether the expected dependency structure of the motif is consistent with statistical tests performed on $D$ and combine the resulting evidence into a motif confidence score. Unlike edge confidence, which evaluates individual causal relationships, motif confidence validates the statistical consistency of the local causal structure as a whole, thereby complementing edge-level validation through higher-order structural consistency among the edges comprising the motif.

\subsection{Edge Resolution}
\label{sec:edge-resolution}
\textbf{Motivation.} A single confidence score is often insufficient to resolve edges with conflicting semantic and statistical evidence. We therefore progressively refine only these uncertain edges, ensuring that every retained edge is supported by increasingly stronger evidence while preserving decision traceability. \\
\textbf{Single score computation.} We combine LLM-derived and data-driven evidence at the level of edge-motif pairs. For each edge $e$ participating in motif $m$, we define score $s$:
{\small
\begin{equation}
\label{eq:score-compute}
s(e,m) = \alpha\, p(m \mid \mathrm{LLM}) 
+ (1-\alpha)\big(\beta_0\, p(e \mid D) + 
\beta_1\, p(m \mid D)\big), 
\end{equation}
} 
where $\alpha \in [0,1]$ controls the trade-off between LLM and data evidences, and $\beta_0, \beta_1$ balance edge- and motif-level data evidence. An initial graph $G$ is constructed by including edges that satisfy:
$ s(e,m) \geq \lambda_{\text{low}} $
for at least one motif $m~\in~\mathcal{M}$. This step prioritizes recall, allowing uncertain edges to be considered for later refinement.
Edges with intermediate score or conflicting evidences between LLM and data are refined in a second stage. Define the uncertain edges set as CF edges, $E_2$:
\[
E_2:=\left\{
e\in G \; \middle| \;
\begin{array}{l}
\text{(i) } s(e,m)\in[\lambda_{\text{low}},\,\lambda_{\text{high}}], \\[0.3em]
\text{\hspace{1.8em}or}\\[0.3em]
\text{(ii) } p(m|\mathrm{LLM}_{\mathrm{mf}})\, \text{and}\,
p(m|D)\, \text{are} \\
\hspace{1.5em}\text{inconsistent for at least one } \\
\hspace{1.5em} \text{motif }m\ni e.
\end{array}
\right\}.
\] \\
\begin{table*}
\centering
\caption{Performance comparison of \textsc{Genesis} with traditional and LLM-assisted baselines at sample size 1000 using the SHD.}
\label{tab:results_1000}
\resizebox{\textwidth}{!}{
\begin{tabular}{lccccccccc}
\hline
Dataset/Model & CausalOrder & LLM-CD & SLLM-CD & PC & SCORE & NOTEARS & CaMML & Direct-LiNGAM & \textsc{Genesis} \\
\hline
Earthquake & \textbf{0.00$\pm$0.00} & \textbf{0.00$\pm$0.00} & 3.16$\pm$0.41 & 3.17$\pm$0.90 & 9.00$\pm$0.63 & 4.00$\pm$0.00 & \underline{1.00$\pm$1.10} & 2.33$\pm$0.51 & \textbf{0.00$\pm$0.00} \\
Cancer & \underline{0.83$\pm$0.75} & 1.33$\pm$0.81 & 1.67$\pm$0.52 & 2.50$\pm$0.50 & 8.00$\pm$0.00 & 4.00$\pm$0.00 & 2.67$\pm$0.52 & 2.67$\pm$0.52 & \textbf{0.17$\pm$0.37} \\
Survey & \underline{3.28$\pm$0.41} & \textbf{4.00$\pm$0.63} & 6.00$\pm$0.00 & 5.67$\pm$2.13 & 11.67$\pm$1.21 & 6.00$\pm$0.00 & 5.00$\pm$0.00 & 4.83$\pm$0.41 & 4.17$\pm$1.34 \\
Asia & \underline{0.66$\pm$0.81} & 1.33$\pm$0.82 & 3.33$\pm$0.51 & 10.83$\pm$0.98 & 21.17$\pm$3.02 & 7.83$\pm$0.41 & 2.67$\pm$1.63 & 3.00$\pm$1.79 & \textbf{0.33$\pm$0.47} \\
Child & 11.33$\pm$2.16 & \textbf{2.83$\pm$1.16} & 24.00$\pm$0.00 & 19.17$\pm$2.27 & 172.33$\pm$3.82 & 20.67$\pm$1.21 & \underline{3.83$\pm$1.23} & 40.00$\pm$3.84 & 6.83$\pm$1.86 \\
\hline
\end{tabular}}
\end{table*}
\begin{table*}
\centering
\caption{Performance comparison of \textsc{Genesis} with traditional and LLM-assisted baselines at sample size 2500 using the SHD.}
\label{tab:results_2500}
\resizebox{\textwidth}{!}{
\begin{tabular}{lccccccccc}
\hline
Dataset/Model & CausalOrder & LLM-CD & SLLM-CD & PC & SCORE & NOTEARS & CaMML & Direct-LiNGAM & \textsc{Genesis} \\
\hline
Earthquake & \textbf{0.00$\pm$0.00} & 0\textbf{.00$\pm$0.00} & 2.00$\pm$0.00 & 3.83$\pm$0.90 & 7.50$\pm$1.22 & 4.00$\pm$0.00 & \underline{0.50$\pm$1.22} & 2.50$\pm$0.55 & \textbf{0.00$\pm$0.00} \\
Cancer & 0.83$\pm$0.98 & 2.83$\pm$1.60 & \underline{0.83$\pm$0.75} & 2.50$\pm$0.50 & 8.00$\pm$1.10 & 4.00$\pm$0.00 & 2.50$\pm$0.55 & 2.50$\pm$0.55 & \textbf{0.33$\pm$0.47} \\
Survey & \underline{3.00$\pm$0.84} & 4.00$\pm$0.63 & 5.17$\pm$0.41 & 5.17$\pm$0.69 & 11.50$\pm$1.38 & 6.00$\pm$0.00 & 4.50$\pm$1.05 & 4.83$\pm$0.75 & \textbf{2.67$\pm$0.75} \\
Asia & \underline{1.00$\pm$0.90} & \textbf{1.00$\pm$0.00} & 3.16$\pm$0.41 & 6.83$\pm$0.90 & 21.33$\pm$2.58 & 8.00$\pm$0.00 & 2.17$\pm$0.41 & 5.33$\pm$3.67 & 1.17$\pm$0.90 \\
Child & 5.67$\pm$0.81 & \textbf{1.67$\pm$1.63} & 24.00$\pm$0.00 & 20.83$\pm$2.67 & 173.33$\pm$3.50 & 20.17$\pm$0.41 & \underline{2.99$\pm$1.10} & 44.00$\pm$4.47 & 6.50$\pm$2.69 \\
\hline
\end{tabular}}
\end{table*}
\textbf{Markov Blanket (MB) Filtering.}
The MB~\cite{tsamardinos2003algorithms} of a node is the minimal set of variables that renders it conditionally independent of all remaining variables, making it an effective mechanism for local structural validation. For each node $a$, we compute a data-derived MB ($\mathrm{MB}(a \mid D)$) using the IAMB algorithm in~\cite{tsamardinos2003algorithms}, and an LLM-derived MB ($\mathrm{MB}(a \mid \mathrm{LLM_{mb}})$), and define their consensus as $
\mathrm{MB}(a)=\mathrm{MB}(a \mid D)\cap \mathrm{MB}(a \mid \mathrm{LLM_{mb}}).$ For each edge $e:a \rightarrow b \in E_2$, the edge is retained if $a \in \mathrm{MB}(b)$ and $b \in \mathrm{MB}(a).$ Otherwise, the edge is deferred to the subsequent stage, given below, for further analysis. \\
\noindent\textbf{Counterfactual LLM Resolution.} Edges that remain unresolved after MB filtering are resolved in two steps. A hypothesis LLM, $\mathrm{LLM}_{\mathrm{hyp}},$ first enumerates five competing structural explanations for the candidate link $e: a \rightarrow b$, a direct edge, two singly-mediated paths, a longer causal chain, and a confounded alternative in which a common parent explains both endpoints, restricted to the closed variable vocabulary induced by $M$ and constrained to first-principles domain reasoning rather than recall of known benchmark structures. The Judge LLM ($\mathrm{LLM}_{\mathrm{judge}}$) is then given $(e, M)$ together with these hypotheses and the statistical evidence
accumulated for $e$, and performs counterfactual reasoning to assess how the plausibility of $e$ changes under each alternative local configuration before
producing a binary decision (KEEP or DISCARD). Because the decision is an adjudication among explicit, enumerated alternatives rather than an answer to
an unconstrained query, the retained hypothesis and the reason for rejecting its competitors are both recoverable, which is what makes the Judge LLM a
valid evidence source under Definition~1. The retained edges are combined with those accepted during MB filtering to construct the final DAG $G$.

Overall, \textsc{Genesis} progressively integrates LLM-derived semantic priors with statistical evidence, resolving structural ambiguities through increasingly stringent validation stages. This progressive refinement ensures that every edge in the final DAG is supported by explicit, auditable evidence, thereby achieving decision traceability by design.
\section{Numerical results}
\label{sec:experiments}
\begin{table*}
\centering
\caption{Performance comparison of \textsc{Genesis} with traditional and LLM-assisted baselines at sample size 5000 using the SHD.}
\label{tab:results_5000}
\resizebox{\textwidth}{!}{
\begin{tabular}{lccccccccc}
\hline
Dataset/Model & CausalOrder & LLM-CD & SLLM-CD & PC & SCORE & NOTEARS & CaMML & Direct-LiNGAM & \textsc{Genesis} \\
\hline
Earthquake & \textbf{0.00$\pm$0.00} & \textbf{0.00$\pm$0.00} & \underline{0.17$\pm$0.41} & 4.00$\pm$0.00 & 8.00$\pm$0.89 & 4.00$\pm$0.00 & \textbf{0.00$\pm$0.00} & 2.33$\pm$0.52 & \textbf{0.00$\pm$0.00} \\
Cancer & \textbf{0.00$\pm$0.00} & \underline{0.67$\pm$0.51} & 2.17$\pm$1.47 & 2.00$\pm$0.00 & 8.00$\pm$0.63 & 4.00$\pm$0.00 & 2.50$\pm$0.55 & 2.00$\pm$0.00 & \textbf{0.00$\pm$0.00} \\
Survey & \textbf{1.50$\pm$0.84} & \underline{1.50$\pm$1.22} & 5.33$\pm$0.51 & 5.67$\pm$0.94 & 11.33$\pm$1.51 & 6.00$\pm$0.00 & 3.83$\pm$1.17 & 4.33$\pm$0.82 & 1.83$\pm$0.90 \\
Asia & \textbf{0.17$\pm$0.41} & \underline{0.67$\pm$0.81} & 3.16$\pm$0.41 & 7.17$\pm$0.37 & 22.17$\pm$2.56 & 8.00$\pm$0.00 & 2.67$\pm$1.63 & 3.83$\pm$2.56 & 1.00$\pm$0.82 \\
Child & 5.67$\pm$0.81 & \textbf{1.00$\pm$0.00} & 24.00$\pm$0.00 & 18.50$\pm$2.29 & 172.00$\pm$4.10 & 20.00$\pm$0.00 & \underline{1.67$\pm$1.03} & 54.50$\pm$4.37 & 6.67$\pm$1.89 \\
\hline
\end{tabular}}
\end{table*}

We evaluate \textsc{Genesis} using two complementary criteria. We first report decision traceability by analyzing the distribution of justification sources across edge decisions, quantifying how statistical evidence, MB consistency, and Judge LLM reasoning contribute to the construction of the final DAG. We then report SHD, the standard metric for structural accuracy, to compare \textsc{Genesis} against existing CD methods on benchmark datasets.\\
\textbf{Datasets} We evaluate \textsc{Genesis} on five standard benchmark Bayesian networks widely used in CD: \textit{Asia}, \textit{Cancer}, \textit{Child}, \textit{Earthquake}, and \textit{Survey}. These datasets span a range of graph sizes and structural complexities. For each benchmark, we generate observational datasets of $1,000$, $2,500$ and $5,000$ samples from the corresponding ground-truth DAG. Hyperparameter settings, decoding parameters, and the full evaluation protocol
are provided in Appendix.
\subsection{Decision Traceability}
\textbf{Traceability Distribution.}
Table~\ref{tab:traceability} reports the distribution of justification sources across edge decisions for each benchmark dataset at $n=1000$. Similar results are observed across all sample sizes; the corresponding results are provided in the Appendix due to space constraints. Across all five benchmark networks, \textsc{Genesis} achieves \emph{100\% decision traceability} by construction, with every retained edge supported by at least one explicit, auditable justification. Most edge decisions are resolved through statistical evidence and MB consistency, while the Judge LLM is invoked only when data-driven evidence is insufficient, serving as a targeted fallback rather than the primary decision mechanism. Further, Table~\ref{tab:trace-cancer} illustrates edge-level decision traceability on the Cancer dataset for one random seed, showing how each discovered edge is attributed to its corresponding justification source and matches the ground truth. \\
\textbf{Illustrative Trace.} We illustrate the decision traceability mechanism using the edge $\textit{Smoker} \rightarrow \textit{Cancer}$ in the Cancer dataset (Figure~\ref{fig:traceability}). The LLM Explorer assigns a prior score $p(m \mid \mathrm{LLM_{mf}})$ of 0.99,  via the collider motif $\{\text{Smoker} \rightarrow \text{Cancer} \leftarrow 
\text{Pollution}\}$. The Data Explorer computes an edge confidence $p(e \mid D)$ of 0.72, and a motif confidence $p(m \mid D)$ of 0.20, yielding a combined score $s(e, m)$ of 0.62, within the uncertainty interval $[\lambda_\text{low} = 0.4, \lambda_\text{high} = 0.7]$. MB validation then confirms that $\text{Smoker} \in \text{MB}(\text{Cancer})$ and $\text{Cancer} \in \text{MB}(\text{Smoker})$, satisfying the structural necessity criterion in Step 3(a) of \textsc{Genesis} given in Algorithm-1. The edge is retained without invoking the Judge LLM, with the decision jointly traceable to statistical edge confidence and MB consistency. Critically, this justification remains meaningful in the absence of a ground-truth graph: a domain expert can independently evaluate the cited statistical and structural evidence without requiring external validation.
\subsection{Structural Accuracy Performance of \textsc{Genesis}}
We evaluate the structural accuracy of \textsc{Genesis} using SHD against SOTA traditional and LLM-based CD methods. \\
\textbf{Traditional Baselines.} 
We compare \textsc{Genesis} against representative constraint-based, score-based, Bayesian, and functional CD methods: PC (conditional independence tests), NOTEARS (continuous optimization with acyclicity constraints), SCORE (order-based causal discovery using score functions), CaMML (minimum message length-based Bayesian learning), and DirectLiNGAM (functional causal models exploiting non-Gaussianity). \\
\noindent\textbf{LLM-based Baselines.} We compare against recent LLM-assisted CD methods: CausalOrder~\cite{vashishtha2025causal}, which infers a global causal
ordering to constrain downstream CD; LLM-CD~\cite{ban2025integrating}, which incorporates LLM-derived causal priors as soft structural constraints; and
SLLM-CD~\cite{du2025causal}, which iteratively combines LLM-based semantic reasoning with statistical evidence for causal structure learning.\\
\textbf{Performance of \textsc{Genesis}.} Tables~\ref{tab:results_1000}--\ref{tab:results_5000} summarize the performance of all methods across varying sample sizes (1000, 2500 and 5000). All results are averaged over six runs with different random seeds, and we report the mean and standard deviation. Overall, \textsc{Genesis} consistently outperforms statistical CD methods on the majority of benchmark datasets across all sample regimes in terms of SHD, while achieving performance comparable to SOTA LLM-assisted approaches. \\
\textbf{Robustness across sample sizes.}
A key observation is that \textsc{Genesis} maintains stable performance as sample size increases. On smaller networks such as \textit{Earthquake} and \textit{Cancer}, our algorithm achieves near-zero SHD across nearly all sample sizes, indicating that motif-level reasoning combined with statistical validation effectively recovers simple causal structures even under limited data. Unlike several baselines, the performance of \textsc{Genesis} does not fluctuate significantly between low- and high-sample regimes. \\
\textbf{Comparison with traditional methods.}
Constraint-based and score-based methods, including PC and NOTEARS, consistently exhibit higher SHD across most datasets. Although larger sample sizes improve their performance, they continue to struggle on datasets such as \textit{Asia} and \textit{Child}, where Markov equivalence and complex dependencies remain challenging. Identifiable functional methods, including SCORE and DirectLiNGAM, also exhibit inconsistent performance across datasets, reflecting the sensitivity of identifiable approaches to their underlying assumptions. Among traditional baselines, CaMML is the strongest, particularly on \textit{Child}, but is less consistent on smaller datasets. In contrast, \textsc{Genesis} achieves a better balance between accuracy and robustness across diverse settings
\\
\textbf{Comparison with LLM-based methods.}
LLM-based methods such as CausalOrder, LLM-CD, and SLLM-CD demonstrate mixed behavior across datasets and sample sizes. While they occasionally achieve strong results on smaller networks, their performance shows limited improvement with increasing data, suggesting that they rely primarily on semantic priors rather than statistical refinement. In contrast, \textsc{Genesis} consistently benefits from additional data through its integration of motif-level priors and data-driven validation. \\
\textbf{Performance on complex datasets.}
On structurally complex datasets such as \textit{Child}, all methods experience increased difficulty, reflected in higher SHD values overall. Nevertheless, \textsc{Genesis} remains competitive across all sample sizes and substantially outperforms several traditional and LLM-based baselines. This indicates that motif-level structural reasoning remains effective even in larger and denser causal graphs.
\subsection{LLM Implementation Details}
GENESIS instantiates each LLM role with a distinct model. Motif extraction ($\mathrm{LLM}_{\mathrm{mf}}$) and causal hypothesis generation use GPT-5~\cite{openai2025gpt5}. MB estimation ($\mathrm{LLM}_{\mathrm{mb}}$) uses the open-weight Qwen3.5-27B model~\cite{qwen2026qwen35}, served locally on a single NVIDIA
L40S GPU (48\,GB), chosen for its favourable cost-accuracy trade-off given that MB queries are issued once per node. Counterfactual edge resolution ($\mathrm{LLM}_{\mathrm{judge}}$) uses Gemini~3~Pro~\cite{deepmind2025gemini3}. Instantiating the stages with heterogeneous models avoids correlated failure modes that arise when a single model both proposes and adjudicates the same structural hypothesis, and strengthens the independence of the evidence sources underlying decision traceability~\cite{panickssery2024llm, li2025preference}. 
The exact prompt templates used at every stage are provided in Appendix.
\subsection{Practical Implication}
A method achieving lower SHD without providing edge-level justifications offers limited practical value in real-world deployments, where ground-truth DAGs are unavailable. In contrast, \textsc{Genesis} produces an auditable evidence trail for every edge decision, transforming CD from an opaque optimization process into transparent, evidence-based reasoning. This capability is particularly valuable in high-stakes domains such as healthcare, policy analysis, and industrial fault diagnosis, where practitioners must trust, verify, and communicate discovered causal relationships.

\section{Conclusion}
We introduced \textsc{Genesis}, a hybrid CD framework that elevates \emph{decision traceability} to a first-class objective. By decomposing structure learning into interpretable motif-level reasoning and progressively resolving uncertain edges through statistical evidence, consensus Markov Blankets, and LLM-based domain reasoning, \textsc{Genesis} guarantees that every edge decision is supported by an explicit, auditable justification. Experiments demonstrate that this interpretability is achieved without sacrificing structural accuracy, outperforming statistical baselines on the majority of benchmark datasets while remaining competitive with SOTA LLM-assisted methods. This work highlights the importance of optimizing causal discovery not only for structural accuracy but also for the transparency and accountability of individual causal decisions.

\bibliography{references}

\appendix
\onecolumn
\section{Appendix}
\section{Prompt Templates}
\label{app:prompts}

We provide the exact prompt templates used by all LLMs ($\mathrm{LLM}_{\text{mf}}, \mathrm{LLM}_{\text{mb}}, \mathrm{LLM}_{\text{hyp}}, \mathrm{LLM}_{\text{judge}}$) used in this work.  All prompts enforce a closed variable vocabulary over nodes metadata $M.$

\subsection{Causal Motif Generation (Asia Network)}
\label{app:prompt-motif}
\begin{promptbox}{LLM Explorer --- Causal Motif Generation}
\textbf{Role.} You are a Bayesian Network Medical Expert. Propose exactly
40 causal motifs (graph fragments) to reconstruct the causal structure
of the \emph{Asia} (Chest Clinic) diagnostic network. 
\textbf{Output a JSON array only.}

\medskip
\textbf{Variable Definitions.}
\begin{itemize}[leftmargin=1.4em,itemsep=1pt,topsep=2pt]
  \item \textbf{Asia} --- Binary. Visit to an endemic area (risk factor).
  \item \textbf{Smoking} --- Binary. Tobacco use history (risk factor).
  \item \textbf{Tuberculosis} --- Binary. Presence of TB infection.
  \item \textbf{LungCancer} --- Binary. Presence of lung malignancy.
  \item \textbf{Bronchitis} --- Binary. Chronic bronchial inflammation.
  \item \textbf{Either} --- Binary. Logical variable (True if TB $\lor$ LungCancer).
  \item \textbf{XRay} --- Binary. Chest X-ray result.
  \item \textbf{Dyspnea} --- Binary. Shortness of breath.
\end{itemize}

\medskip
\textbf{Causal Design Patterns.}
\begin{enumerate}[leftmargin=1.6em,itemsep=1pt,topsep=2pt]
  \item \textbf{Epidemiological Risk} (Environment $\to$ Pathology).
  \item \textbf{Logical Synthesis} (Pathology$\to$Aggregation): TB and Lung
        Cancer aggregate in \texttt{Either}.
  \item \textbf{Imaging Causality} (State $\to$ Image).
  \item \textbf{Respiratory Mechanics} (Pathology $\to$ Symptom).
  \item \textbf{Common Cause} (Habit $\to$ Multiple Pathologies).
\end{enumerate}

\textbf{Motif Distribution (40 total).} ${\approx}33\%$ chains
($A\!\to\!B\!\to\!C$), ${\approx}33\%$ forks ($A\!\to\!B,\,A\!\to\!C$),
${\approx}33\%$ colliders ($A\!\to\!C\!\leftarrow\!B$).

\medskip
\textbf{Output Schema (exactly 40 entries).} Each object includes
\texttt{type}, \texttt{vars}, \texttt{implied\_edges}, \texttt{prior} (float),
and \texttt{confidence\_reason}. Output JSON only.
\end{promptbox}

\noindent\textbf{\textcolor{cobaltdark}{Sample Output Entry}}
\begin{lstlisting}[style=jsonstyle]
{  "type"              : "collider",
  "vars"              : ["Tuberculosis", "LungCancer", "Either"],
  "implied_edges"     : ["Tuberculosis->Either", "LungCancer->Either"],
  "prior"             : 1.0,
  "confidence_reason" : "Both TB and Lung Cancer independently produce thoracic
                         pathology sufficient to activate the Either node."}
\end{lstlisting}

\subsection{Markov Blanket Estimation}
\label{app:prompt-mb}

\begin{promptbox}{$\mathrm{LLM_{mb}}$ --- Markov Blanket Estimation \quad\normalfont\small(Sample node: \texttt{Smoking})}
\textbf{Role.} You are a Bayesian Network Medical Expert. Identify the Markov
Blanket of the target node using only the closed variable vocabulary supplied
below. \textbf{Output a JSON object only.}

\medskip
\textbf{Target.} \texttt{Smoking} --- Binary. Tobacco use history (risk factor).

\medskip
\textbf{Other Variables.}
\begin{itemize}[leftmargin=1.4em,itemsep=1pt,topsep=2pt]
  \item \textbf{Asia} --- Binary. Visit to an endemic area (risk factor).
  \item \textbf{Tuberculosis} --- Binary. Presence of TB infection.
  \item \textbf{LungCancer} --- Binary. Presence of lung malignancy.
  \item \textbf{Bronchitis} --- Binary. Chronic bronchial inflammation.
  \item \textbf{Either} --- Binary. Logical variable (True if TB $\lor$ LungCancer).
  \item \textbf{XRay} --- Binary. Chest X-ray result.
  \item \textbf{Dyspnea} --- Binary. Shortness of breath.
\end{itemize}

\textbf{Task.} List the Markov Blanket of \texttt{Smoking}: parents, children,
and co-parents only, restricted to direct relationships. Return an empty array
for any category with no members. Respond only with the JSON object below, with
no explanation.
\end{promptbox}

\noindent\textbf{\textcolor{cobaltdark}{Sample Output Object}}
\begin{lstlisting}[style=jsonstyle]
{
  "target_node"    : "Smoking",
  "parents"        : [],
  "children"       : ["LungCancer", "Bronchitis"],
  "co_parents"     : [],
  "markov_blanket" : ["LungCancer", "Bronchitis"]
}
\end{lstlisting}

\subsection{Hypothesis Generation LLM}
\label{app:prompt-hyp}

\begin{promptbox}{Judge LLM --- Hypothesis Generation \quad\normalfont\small(Sample edge: \texttt{Smoking}\,$\to$\,\texttt{Dyspnea})}
\textbf{Role.} Expert Pulmonologist and Diagnostician. Derive all relationships
from \textbf{first principles only} --- do not rely on memorized benchmark
structures.

\medskip
\textbf{Source:} \texttt{Smoking} (risk factor). \quad
\textbf{Target:} \texttt{Dyspnea}.\quad
\textbf{Vocabulary (closed):} \texttt{\{Asia, Smoking, Tuberculosis, LungCancer,
Bronchitis, Either, XRay, Dyspnea\}}.

\medskip
\textbf{Task --- 5 distinct causal hypotheses:}
\begin{enumerate}[label=\textbf{\Alph*.},leftmargin=1.8em,itemsep=2pt,topsep=2pt]
  \item \textbf{Direct:} \texttt{Smoking}\,$\to$\,\texttt{Dyspnea} --- airway irritation.
  \item \textbf{Mediated:} \texttt{Smoking}\,$\to$\,\texttt{Bronchitis}\,$\to$\,\texttt{Dyspnea}.
  \item \textbf{Mediated:} \texttt{Smoking}\,$\to$\,\texttt{LungCancer}\,$\to$\,\texttt{Dyspnea}.
  \item \textbf{Complex chain:} \texttt{Smoking}\,$\to$\,\texttt{LungCancer}\,$\to$\,\texttt{Either}\,$\to$\,\texttt{Dyspnea}.
  \item \textbf{Spurious:} \texttt{Asia}\,$\to$\,\texttt{Smoking}\,\&\,\texttt{Dyspnea}.
\end{enumerate}
\end{promptbox}

\subsection{Edge Decision (Judge LLM)}
\label{app:prompt-score}

\begin{promptbox}{Judge LLM --- Edge Decision \quad\normalfont\small(Sample edge: \texttt{Either}\,$\to$\,\texttt{XRay})}
\textbf{Role.} Principal Investigator (Pulmonary Medicine $+$ Bayesian
Epidemiology). Evaluate \texttt{Either->XRay} by weighing hypotheses against
statistical evidence.

\medskip
\textbf{Hypotheses.} (A)~Direct: \texttt{Either}\,$\to$\,\texttt{XRay};
(B)~Mediated: \texttt{Either}\,$\to$\,\texttt{Dyspnea}\,$\to$\,\texttt{XRay};
(C)~Spurious: \texttt{Smoking}\,$\to$\,\texttt{Either}\,\&\,\texttt{XRay}.

\medskip
\textbf{Statistical evidence.} Frequency $=0.93$; MI $=0.181$; bootstrap
confidence $=94\%$ (94/100 samples).

\medskip
\textbf{Decision algorithm.}
(1)~confidence $>80\%$ $\Rightarrow$ real dependency unless a confounder fully
explains it;
(2)~confounder override if a common parent explains it better;
(3)~confidence $<50\%$ with tenuous mechanism $\Rightarrow$ remove.
\end{promptbox}

\noindent\textbf{\textcolor{cobaltdark}{Sample Decision Object}}
\begin{lstlisting}[style=jsonstyle]
{
  "edge"             : "Either->XRay",
  "decision"         : "KEEP",
  "confidence_score" : 96,
  "justification"    : "Statistically dominant direct link grounded in imaging
                        physics; no viable confounder (Hypothesis C rejected)."
}
\end{lstlisting}

\section{Experimental Setup and Hyperparameters}
\label{app:implementation}

\subsection*{Model Instantiation}
GENESIS instantiates each LLM role with a distinct model. Motif extraction
($\mathrm{LLM}_{\mathrm{mf}}$) and causal hypothesis generation
($\mathrm{LLM}_{\mathrm{hyp}}$) use GPT-5~\cite{openai2025gpt5}. Markov Blanket
estimation ($\mathrm{LLM}_{\mathrm{mb}}$) uses the open-weight
Qwen3.5-27B model~\cite{qwen2026qwen35}, served locally on a single NVIDIA
L40S GPU (48\,GB), chosen for its favorable cost--accuracy trade-off given
that MB queries are issued once per node. Counterfactual edge resolution
($\mathrm{LLM}_{\mathrm{judge}}$) uses Gemini 3 Pro~\cite{deepmind2025gemini3}.
Instantiating the stages with heterogeneous models mitigates correlated failure
modes that arise when a single model both proposes and adjudicates the same
structural hypothesis, strengthening the independence of the evidence sources
underlying decision traceability.

\subsection*{Hyperparameters}
Table~\ref{tab:hyperparams} lists all hyperparameters used in Algorithm~1. The
same values are used across all five benchmark networks and all three sample
sizes; no per-dataset tuning is performed.

\begin{table}[h]
\centering
\small
\begin{tabular}{llc}
\toprule
\textbf{Symbol} & \textbf{Description} & \textbf{Value} \\
\midrule

\multicolumn{3}{l}{\emph{Data Explorer --- edge confidence (Eq.~1)}} \\
$w_1$ & Weight, conditional independence (CI) & 0.50 \\
$w_2$ & Weight, BIC score & 0.20 \\
$w_3$ & Weight, additive noise model (ANM) & 0.15 \\
$w_4$ & Weight, LiNGAM & 0.15 \\

\midrule
\multicolumn{3}{l}{\emph{Edge Resolution --- combined score (Eq.~2)}} \\
$\alpha$ & LLM vs.\ data trade-off & 0.25 \\ 
$\beta_0$ & Weight, edge-level data evidence & 0.40 \\
$\beta_1$ & Weight, motif-level data evidence & 0.60 \\
$\lambda_{\mathrm{low}}$ & Lower inclusion threshold & 0.25 \\
$\lambda_{\mathrm{high}}$ & Upper certainty threshold & 0.70 \\

\midrule
\multicolumn{3}{l}{\emph{Motif and hypothesis generation}} \\
$|\mathcal{M}|_{\max}$ & Max.\ motifs extracted & $40$ \\
$|\mathcal{H}(e)|$ & Hypotheses per unresolved edge & 5 \\

\midrule
\multicolumn{3}{l}{\emph{Statistical tests}} \\
$\alpha_{\mathrm{CI}}$ & Significance level, CI tests & 0.05 \\
--- & CI test used & OLS regression \\
--- & Bootstrap resamples & 5 \\

\midrule
\multicolumn{3}{l}{\emph{Markov Blanket estimation}} \\
--- & Data-derived MB algorithm & IAMB \\
$\alpha_{\mathrm{MB}}$ & Significance level, IAMB & 0.05 \\

\bottomrule
\end{tabular}
\caption{Hyperparameter settings used across all experiments.}
\label{tab:hyperparams}
\end{table}

\subsection*{Evaluation Protocol}
All results are averaged over six runs with different random seeds $(0, 1, 2, 3, 4, 5)$;
we report mean and standard deviation. Observational datasets of $1{,}000$,
$2{,}500$ and $5{,}000$ samples are drawn from each ground-truth DAG
settings.

\section{Edge-Level Traceability Analysis ($n=1000$)}
\label{app:traceability}
For each dataset we report every retained edge, the evidence source that
resolved it, and whether the decision matches the ground-truth DAG. Sources are
\src{Stat} (data-driven confidence scoring), \src{MB} (Markov Blanket validation
on the intersection of data- and LLM-derived neighborhoods), and \src{Judge}
(counterfactual-style LLM arbitration). Per-source accuracy is summarized beneath
each table.

\newcolumntype{L}{>{\raggedright\arraybackslash}p{0.50\linewidth}}

\begin{table}[h]
\centering
\renewcommand{\arraystretch}{1.15}
\setlength{\tabcolsep}{8pt}
\begin{tabular}{@{}L l c@{}}
\toprule
\textbf{Edge} & \textbf{Resolved by} & \textbf{Correct} \\
\midrule
LungParench $\to$ HypoxiaInO2   & \src{MB}    & \yes \\
LungFlow $\to$ Sick             & \src{Stat}  & \no  \\
CardiacMixing $\to$ HypDistrib  & \src{MB}    & \yes \\
LungFlow $\to$ ChestXray        & \src{MB}    & \yes \\
HypDistrib $\to$ HypoxiaInO2    & \src{Stat}  & \no  \\
Disease $\to$ CardiacMixing     & \src{MB}    & \yes \\
Disease $\to$ LungParench       & \src{Judge} & \yes \\
ChestXray $\to$ XrayReport      & \src{MB}    & \yes \\
Grunting $\to$ GruntingReport   & \src{MB}    & \yes \\
HypoxiaInO2 $\to$ RUQO2         & \src{MB}    & \yes \\
HypoxiaInO2 $\to$ LowerBodyO2   & \src{MB}    & \yes \\
DuctFlow $\to$ LungFlow         & \src{Stat}  & \no  \\
Disease $\to$ DuctFlow          & \src{MB}    & \yes \\
Disease $\to$ LungFlow          & \src{MB}    & \yes \\
DuctFlow $\to$ CardiacMixing    & \src{Stat}  & \no  \\
Disease $\to$ LVH               & \src{Judge} & \yes \\
LungParench $\to$ LungFlow      & \src{MB}    & \no  \\
Disease $\to$ Age               & \src{Stat}  & \yes \\
Sick $\to$ Age                  & \src{MB}    & \yes \\
LVH $\to$ LVHreport             & \src{Stat}  & \yes \\
CO2 $\to$ CO2Report             & \src{MB}    & \yes \\
Sick $\to$ Grunting             & \src{Stat}  & \yes \\
Disease $\to$ Sick              & \src{Stat}  & \yes \\
CardiacMixing $\to$ HypoxiaInO2 & \src{Stat}  & \yes \\
LungFlow $\to$ HypoxiaInO2      & \src{Stat}  & \no  \\
DuctFlow $\to$ HypDistrib       & \src{Stat}  & \yes \\
\bottomrule
\end{tabular}
\caption{\textbf{Child} (26 edges). Per-source accuracy:
\src{Stat} 6/11 ($55\%$), \src{MB} 11/12 ($92\%$), \src{Judge} 3/3 ($100\%$).}
\label{tab:trace-child}
\end{table}

\begin{table}[h]
\centering
\renewcommand{\arraystretch}{1.15}
\setlength{\tabcolsep}{8pt}
\begin{tabular}{@{}L l c@{}}
\toprule
\textbf{Edge} & \textbf{Resolved by} & \textbf{Correct} \\
\midrule
Tuberculosis $\to$ Either  & \src{MB}    & \yes \\
LungCancer $\to$ Either    & \src{MB}    & \yes \\
Bronchitis $\to$ Dyspnea   & \src{MB}    & \yes \\
Either $\to$ XRay          & \src{MB}    & \yes \\
Smoking $\to$ Bronchitis   & \src{MB}    & \yes \\
Either $\to$ Dyspnea       & \src{Judge} & \yes \\
LungCancer $\to$ Dyspnea   & \src{MB}    & \no  \\
Tuberculosis $\to$ Dyspnea & \src{MB}    & \no  \\
Smoking $\to$ LungCancer   & \src{Judge} & \yes \\
\bottomrule
\end{tabular}
\caption{\textbf{Asia} (9 edges). Per-source accuracy:
\src{MB} 5/7 ($71\%$), \src{Judge} 2/2 ($100\%$).}
\label{tab:trace-asia}
\end{table}

\begin{table}[h]
\centering
\renewcommand{\arraystretch}{1.15}
\setlength{\tabcolsep}{8pt}
\begin{tabular}{@{}L l c@{}}
\toprule
\textbf{Edge} & \textbf{Resolved by} & \textbf{Correct} \\
\midrule
Cancer $\to$ Xray      & \src{MB}   & \yes \\
Pollution $\to$ Cancer & \src{MB}   & \yes \\
Smoker $\to$ Cancer    & \src{MB}   & \yes \\
Cancer $\to$ Dyspnoea  & \src{Stat} & \yes \\
\bottomrule
\end{tabular}
\caption{\textbf{Cancer} (4 edges). Per-source accuracy:
\src{Stat} 1/1 ($100\%$), \src{MB} 3/3 ($100\%$).}
\label{tab:trace-cancer2}
\end{table}

\begin{table}[h]
\centering
\renewcommand{\arraystretch}{1.15}
\setlength{\tabcolsep}{8pt}
\begin{tabular}{@{}L l c@{}}
\toprule
\textbf{Edge} & \textbf{Resolved by} & \textbf{Correct} \\
\midrule
Alarm $\to$ JohnCalls   & \src{MB} & \yes \\
Alarm $\to$ MaryCalls   & \src{MB} & \yes \\
Earthquake $\to$ Alarm  & \src{MB} & \yes \\
Burglary $\to$ Alarm    & \src{MB} & \yes \\
\bottomrule
\end{tabular}
\caption{\textbf{Earthquake} (4 edges). Per-source accuracy: \src{MB} 4/4 ($100\%$).}
\label{tab:trace-earthquake}
\end{table}

\begin{table}[h]
\centering
\renewcommand{\arraystretch}{1.15}
\setlength{\tabcolsep}{8pt}
\begin{tabular}{@{}L l c@{}}
\toprule
\textbf{Edge} & \textbf{Resolved by} & \textbf{Correct} \\
\midrule
Education $\to$ Occupation & \src{MB} & \yes \\
Education $\to$ Residence  & \src{MB} & \yes \\
Age $\to$ Education        & \src{MB} & \yes \\
Residence $\to$ Travel     & \src{MB} & \yes \\
\bottomrule
\end{tabular}
\caption{\textbf{Survey} (4 edges). Per-source accuracy: \src{MB} 4/4 ($100\%$).}
\label{tab:trace-survey}
\end{table}

\begin{table}[h]
\centering
\caption{Distribution of decision justification sources across datasets at $n=2500$. Each value reports the percentage of edges in the final DAG whose decision is supported by the listed evidence source. All edges achieve 
decision traceability by construction.}
\begin{tabular}{lcccc}
\toprule
\textbf{Dataset} & \textbf{Statistical} & 
\textbf{MB Consistency} & \textbf{Judge LLM} & 
\textbf{Decision Traceability} \\
\midrule
Earthquake & 0.00\%  & 100.00\% & 0.00\%  & 100\% \\
Cancer     & 25.00\% & 75.00\%  & 0.00\%  & 100\% \\
Survey     & 0.00\%  & 85.71\% & 14.29\%  & 100\% \\
Asia       & 0.00\%  & 100.00\%  & 0.00\% & 100\% \\
Child      & 9.38\% & 59.38\%  & 31.25\% & 100\% \\
\bottomrule
\end{tabular}
\label{tab:traceability-2500}
\end{table}

\begin{table}[h]
\centering
\caption{Distribution of decision justification sources across datasets at $n=5000$. Each value reports the percentage of edges in the final DAG whose decision is supported by the listed evidence source. All edges achieve 
decision traceability by construction.}
\begin{tabular}{lcccc}
\toprule
\textbf{Dataset} & \textbf{Statistical} & 
\textbf{MB Consistency} & \textbf{Judge LLM} & 
\textbf{Decision Traceability} \\
\midrule
Earthquake & 0.00\%  & 100.00\% & 0.00\%  & 100\% \\
Cancer     & 25.00\% & 75.00\%  & 0.00\%  & 100\% \\
Survey     & 0.00\%  & 100.00\% & 0.00\%  & 100\% \\
Asia       & 0.00\%  & 100.00\%  & 0.00\% & 100\% \\
Child      & 9.38\% & 59.38\%  & 31.25\% & 100\% \\
\bottomrule
\end{tabular}
\label{tab:traceability-5000}
\end{table}

\end{document}